\pdfoutput=1
\documentclass[10pt,twocolumn,letterpaper]{article}
\usepackage{cvpr}              

\makeatletter
\@namedef{ver@everyshi.sty}{}
\makeatother

\usepackage{subcaption}
\usepackage{amsmath}
\usepackage{amssymb}
\usepackage{booktabs}
\usepackage{todonotes}
\usepackage{dsfont}
\usepackage{subcaption}
\usepackage{amsmath}
\usepackage{graphics}
\usepackage{arydshln}
\usepackage{enumitem}
\usepackage[utf8]{inputenc}
\usepackage{placeins}
\usepackage[accsupp]{axessibility}  

\usepackage{tikz}
\usetikzlibrary{fit}
\usepackage{pgfplots}
\pgfplotsset{compat=1.17}

\usepackage[pagebackref,breaklinks,colorlinks]{hyperref}

\usepackage[capitalize]{cleveref}
\crefname{section}{Sec.}{Secs.}
\Crefname{section}{Section}{Sections}
\Crefname{table}{Table}{Tables}
\crefname{table}{Tab.}{Tabs.}

\def\cvprPaperID{16} 
\def\confName{XAI4CV}
\def\confYear{2023}

\begin{document}

\title{Localized Shortcut Removal}

\author{
Nicolas M. Müller \thanks{equal contribution}  \\
Fraunhofer AISEC, Germany \\
{\tt\small nicolas.mueller@aisec.fraunhofer.de}
\and
Jochen Jacobs \footnotemark[1] \\
TU Munich, Germany \\
{\tt\small jochen.jacobs@tum.de}
\and
Jennifer Williams \\
University of Southampton, UK \\
{\tt\small J.Williams@soton.ac.uk}
\and
Konstantin Böttinger \\
Fraunhofer AISEC, Germany \\
{\tt\small konstantin.boettinger@aisec.fraunhofer.de}
}
\maketitle

\begin{abstract}
Machine learning is a data-driven field, and the quality of the underlying datasets plays a crucial role in learning success. However, high performance on held-out test data does not necessarily indicate that a model generalizes or learns anything meaningful. This is often due to the existence of machine learning shortcuts - features in the data that are predictive but unrelated to the problem at hand.
To address this issue for datasets where the shortcuts are smaller and more localized than true features, we propose a novel approach to detect and remove them. We use an adversarially trained lens to detect and eliminate highly predictive but semantically unconnected clues in images. 
In our experiments on both synthetic and real-world data, we show that our proposed approach reliably identifies and neutralizes such shortcuts without causing degradation of model performance on clean data. We believe that our approach can lead to more meaningful and generalizable machine learning models, especially in scenarios where the quality of the underlying datasets is crucial.
\end{abstract}

\section{Introduction}

Shortcuts in machine learning data refer to false features that are strongly correlated with the target class but are not expected to be present in real-world applications. These features are easy for neural networks to learn, but they may not generalize beyond the training data. Shortcuts can arise from various factors, such as the data collection process, data collection techniques, or the type of data being collected.
Often, these shortcuts are highly localized and spatially much smaller than true features~\cite{lapuschkin2019unmasking,Zech2018,degrave2021ai}. For instance, a neural network trained on an image dataset where all images of class $k$ exclusively contain watermarks has been shown to rely solely on the presence of the watermark to predict the class \cite{ThePASCA29:online, lapuschkin2019unmasking}.
Indeed, identifying shortcuts during data collection or preprocessing can be a challenging task. This is evidenced by the fact that there are many datasets released to the public that contain shortcuts\cite{ThePASCA29:online,Mueller2021,degrave2021ai,Tahir2021}.

Training a model on data with shortcuts can lead to an over-reliance on irrelevant features.
This results in seemingly high performance on held-out data if the shortcut is present, which may be the case if the test data is sampled via the same process as the training data. 
Such models may not generalize well to out-of-distribution (OOD) data, which is a common issue in machine learning known as domain generalization \cite{Zhou2022}.



In this paper, we introduce a supervised neural network that can learn the essential features of a dataset, even if there are localized shortcuts present (known or unknown). To accomplish this, we use an adversarially trained ``neural lens'' that can remove shortcut features and provide a visual representation of the avoided shortcuts.
Our model is successful in identifying and in-painting shortcuts in various datasets, such as chest x-rays from the COVID QU-Ex dataset~\cite{Tahir2021}. Importantly, this process doesn't harm the model's performance when no shortcuts are present.


\section{Related Work}
In machine learning, shortcuts come in varying degrees of spatiality, ranging from small and localized to global. Local examples include logos and watermarks in image datasets, such as the Pascal VOC 2007 dataset's watermark on horse photos \cite{ThePASCA29:online, lapuschkin2019unmasking}, or hospital- or device-specific marks in chest x-ray images \cite{Zech2018, degrave2021ai}. Meanwhile, global shortcuts include the presence of pastures as an easy indicator for the class "Cow" \cite{Beery2018}, or artefacts in pooled medical databases, where patient positioning, imaging device type, and image size are utilized by the model to infer the target class \cite{robinson2021deep}. 
These shortcuts are problematic not only in supervised computer vision but also in self-supervised learning \cite{Gidaris2018} and when using pretext tasks to design feature extractors \cite{Doersch2015, Minderer2020}. Additionally, shortcuts are not limited to image datasets; they can also be observed in audio datasets. For instance, the amount of leading silence in the ASVspoof Challenge Dataset on audio deepfake detection can be utilized to predict the target class \cite{wang2020asvspoof, Mueller2021}.

\subsection{Automatic shortcut removal}

One potential solution to address the presence of shortcuts in a dataset is to remove them. For instance, in the context of self-supervised representation learning, Minderer et al. \cite{Minderer2020} suggest incorporating a U-Net \cite{Ronneberger2015}, referred to as a "lens," in front of the classification network. The lens is trained adversarially and enables the elimination of local shortcuts, such as logos, through in-painting. However, this approach is restricted to self-supervised learning.
In the supervised domain, adversarial autoencoders have been proposed by Baluja et al. \cite{Baluja2017} and Poursaeed et al. \cite{Poursaeed2018}. 
In this approach, an autoencoder is added at the beginning of a classification network and trained adversarially to generate images that appear similar to the original input, but can mislead the classifier into producing incorrect output. 
Similarly, Xiao et al. \cite{Xiao2019} introduce AdvGAN, which incorporates a GAN-discriminator as an additional loss for the autoencoder, leading to less noticeable perturbations. While these methods share similarities with the architecture proposed in this work, none utilize the generated adversarial images to robustly train the classifier.

\subsection{Improving model robustness}


An alternative approach for addressing shortcuts is to enhance the robustness of models against them. Wang et al. \cite{wang2019learning} propose 
the use of gradient-reversal to deceive helper networks that consider only small local patches, while the global network is encouraged to classify the overall input correctly. A similar idea is explored in \cite{dagaev2023too}.
To prevent a network from focusing excessively on shortcuts that exist only in a subset of the dataset, Dagaev et al. \cite{Dagaev2021} suggest weighted training, which involves assigning lower weights to images that can be accurately classified by a low-capacity network, assuming that those contain shortcuts. However, this approach may not be effective when a significant number of images in the dataset contain shortcuts, unlike our proposed method, c.f. \cref{ss:eval_synth}.
Lastly, for known shortcuts, one can artificially introduce them into the dataset and encourage the model to disregard them \cite{bahng2020learning}. The drawback of this approach is that the shortcuts must be identified beforehand.

\section{Architecture}
\label{sec:architecture}
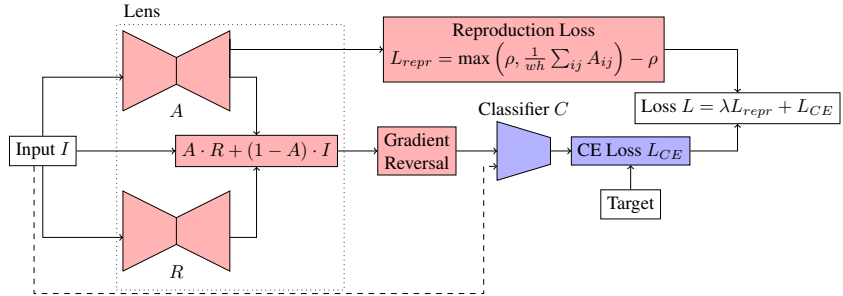
\begin{figure*}
    \centering
    \tikzset{
    unet/.pic={
      \path[draw=black, fill=red!30] (-1,-0.75) -- (0,-0.25) -- (1,-0.75) -- (1,0.75) -- (0,0.25) -- (-1,0.75) -- (-1,-0.75);
      \draw (0,-0.25) -- (0,0.25);
      \node at (0,-0.75) {#1};
    },
    clsf/.pic={
      \path[draw=black, fill=blue!30] (-1,-0.55) -- (0,-0.25) -- (0,0.25) -- (-1,0.55) -- (-1,-0.55);
    },
}

\resizebox{0.64\textwidth}{!}{
\begin{tikzpicture}
    \node [rectangle,draw] (input) at (-1,0) {Input $I$};
    \pic [local bounding box=col] at (1.5,-1.5) {unet=$R$};
    \pic [local bounding box=att] at (1.5,1.5) {unet=$A$};
    \draw [-to] (input) |- (col);
    \draw [-to] (input) |- (att);
    \node [rectangle,draw,fill=red!30] (m1) at (3,0) {$A \cdot R + (1-A) \cdot I $};
    \draw [-to] (att) -| (m1);
    \draw [-to] (col) -| (m1);
    \draw [-to] (input) -- (m1);
    \node [rectangle,draw, align=center, fill=red!30] (gr) at (6,0) {Gradient\\Reversal};
    \draw [-to] (m1) -- (gr);
    \pic [local bounding box=clsf] at (8.5,0) {clsf};
    \node at (8,0.8) {Classifier $C$};
    \draw [-to] (gr) -- (clsf);
    \draw [-to, dashed] (input.240) |- (7.25,-2.67) |- (clsf.210);
    \node [rectangle,draw,] (target) at (10,-1) {Target};
    \node [rectangle,draw,fill=blue!30] (cel) at (10,0) {CE Loss $L_{CE}$};
    \draw [-to] (clsf) -- (cel);
    \draw [-to] (target) -- (cel);
    \node [rectangle,draw, align=center, fill=red!30] (al) at (8,1.9) {Reproduction Loss\\$L_{repr} = \text{max}\left(\rho, \frac{1}{wh} \sum_{ij} A_{ij}\right) - \rho$};
    \draw [-to] (att) -| ( 2.5, 2.1) |- (al);
    \node [rectangle,draw] (loss) at (12,0.8) {Loss $L = \lambda L_{repr} + L_{CE}$};
    \draw [-to] (al) -| (loss);
    \draw [-to] (cel) -| (loss);
    
    \node[draw,dotted,label={[shift={(6ex,0ex)},align=left]north west:Lens}, fit=(att) (col) (m1)] {};

\end{tikzpicture}
}
    \caption{
        Architecture of our proposed \emph{attention lens} model. The lens (red) consists of an attention module $A$ and a reconstruction module $R$, both of which are U-Nets. 
        Its output is passed to the original classifier $C$ (blue), trained via cross-entropy loss. 
        Optionally, input images are also passed to the original classifier. 
        The lens is trained via the classifier's inverted gradients and a reproduction penalty loss $L_{repr}$.
    }
    \label{fig:architecture}
\end{figure*}

To remove shortcuts in supervised problems, we adopt an unsupervised learning architecture ~\cite{Minderer2020}. 
A low-capacity Image-to-Image network (called ``Lens Network'') is placed in front of the classification network.
This lens is then trained jointly, but adversarially, with the classifier to decrease its performance.
The idea is that the lens is trained to isolate features of the image that the classification network is paying attention to.
Since the capacity of the lens is limited, only simple features (i.e. shortcuts) can be removed by the lens. 
To further enforce this, we extend the training loss with an additional reproduction loss $L_{repr}$.
This ensures that the lens modifies the original image only slightly. 

Inspired by \cite{Pumarola2018}, we propose using two U-Net-based networks, an attention network $A$ and a replacement network $R$, as shown in \cref{fig:architecture}. Network $A$ determines the location of the shortcut in the original image, while network $R$ computes a suitable replacement for the shortcut.

Given an input image $I$, we obtain a shortcut-removed image $I'$ as follows:
\begin{equation}
    I' = A \cdot R + (1-A) \cdot I.
\end{equation}
The capacity of the attention network corresponds to the complexity of the shortcuts identified and should be chosen accordingly.
Since the task of the replacement network is more complex than that of the attention network, therefore we accord a larger model capacity (i.e. more up- and downsampling steps) to $R$ than to $A$.
Lens and classification model are trained jointly via 
$L= \lambda L_{repr} + L_{CE}$
where $L_{CE}$ is the cross entropy loss of the classification network $C$ and $\lambda$ is a hyperparameter controlling how much the lens is allowed to modify the input image:
\begin{equation}
    L_{repr} = \text{max}\left(\rho, \frac{1}{wh} \sum_{ij} A_{ij}\right) - \rho.
\end{equation}
$\rho \in [0, 100\%]$ is a hinge hyperparameter that controls the percentage of the image that can be modified without penalty.
Note that while gradients from the cross-entropy loss flow into both the lens and classifier, the reproduction loss only affects the lens.
In our experiments, we use the ResNet18~\cite{He2016} architecture as classifier $C$.

\label{subsec:model_oscillations}
We have noticed oscillations during training, where the classifier stops paying attention to the shortcuts once they are removed, leading the lens to stop removing them. To counteract this, we pass a copy of the image directly to the classifier.
This ensures consistent focus on the shortcuts and attenuates oscillations during training.

\section{Data and synthetic shortcuts}
\label{sec:datasets}
We assess the performance of our proposed architecture on both synthetic and real-world datasets. Initially, we examine our model's efficacy by introducing artificial shortcuts on CIFAR10~\cite{Krizhevsky2009} and ImageNet~\cite{Russakovsky2015}. Specifically, for CIFAR10, we create "Color Dot" and "Location Dot" shortcuts by in-painting a circle in which the color or location corresponds to the target class, as shown in~\cref{fig:cifar_result}. In addition, we use a subset of the visually similar target classes, "goose" and "pelican," from the ImageNet dataset~\cite{Russakovsky2015} to simulate real-world scenarios where classes have overlapping visual features, such as in medical image analysis. To enhance visual similarity, we convert these images to grayscale and introduce shortcuts by overlaying a single logo or a textual watermark across the entire image.

Furthermore, we conduct an evaluation on real-world data, specifically, the covid-qu-Ex dataset \cite{Tahir2021}, which comprises x-ray images of the human chest labelled as either "healthy", "COVID-19", or "pneumonia". Chest X-ray images have been previously found to contain shortcuts \cite{Zech2018}, especially when obtained from multiple sources, such as different hospitals. Upon visually examining the dataset, we observe a significant amount of text, markers, and medical equipment in the corners of the images that may serve as shortcuts, provided they are correlated to the target class. Such shortcuts can severely impede the practical applicability of machine learning models in real-world scenarios~\cite{degrave2021ai}.

\section{Experiments and Results}
\subsection{Synthetic Data}\label{ss:eval_synth}
\begin{table}[t]
    \centering
         \begin{tabular}{|rl|cc|}
    \hline
            & Shortcut       & W/o Lens               & With Lens  \\
    \hline

    CIFAR10 & None           & $\mathbf{75.1 \pm 2.4}$ & $\mathbf{76.7 \pm 2.3}$ \\
            & Color Dot      & $28.5 \pm 0.9$          & $\mathbf{70.5 \pm 2.1}$ \\
            & Location Dot   & $41.9 \pm 7.0$          & $\mathbf{69.0 \pm 3.2}$ \\

    \hline
    ImageNet & None          & $\mathbf{78.9 \pm 1.1}$& $\mathbf{76.1 \pm 2.8}$\\
             & Logo          & $51.9 \pm 2.0$ &  $\mathbf{74.1 \pm 9.0}$ \\
             & Watermark     & $52.4 \pm 1.4$ &  $\mathbf{61.0 \pm 5.2}$ \\

    \hline
\end{tabular}
    \caption{The effect of the lens network, measured in test accuracy. We train a ResNet18 architecture on datasets with and without shortcuts and subsequently assess the model's performance on clean validation data. The experiment is repeated three times, and the mean test accuracy and a $95\%$ confidence interval are reported.
    }
    \label{tab:results}
\end{table}
This section presents the results of our proposed model when training on shortcut-perturbed data, and evaluating on clean test data (CIFAR and ImageNet).

\textbf{Experimental Setup}. For the attention network $A$, we chose 3~downsampling steps and 5~downsampling steps for the replacement network $R$. 
For the CIFAR-based experiments, we use $\rho=2.5\%$, while for the ImageNet experiments, we use $\rho=5.0\%$ (logo shortcut) or $\rho=10.0\%$ (watermark shortcut).
Classifier and Lense have different learning rate ($1.5\cdot10^{-6}$ and $1 \cdot 10^{-4}$, respectively).
We use $\lambda=15$ and train the model for $30$ epochs on CIFAR10, and $50$ epochs on ImageNet.

\textbf{Results}.
Based on the results presented in \Cref{tab:results}, we make the following observations. Firstly, the absence of shortcuts does not impair the test accuracy, indicating that our proposed solution is effective without any drawbacks. Secondly, our proposed shortcuts prove to be highly effective, leading to a substantial decrease in test performance (first row). For instance, the Color Dot shortcut lowers the accuracy from $75\%$ to $28.5\%$, reflecting the model's over-reliance on the simplistic shortcut features. However, with the lens activated, the adverse impact of the shortcuts is almost entirely mitigated. The performance of the "Color Dot" shortcut on CIFAR10 is restored from $28.5\%$ to $70.5\%$ of the original $75\%$, for example.

\begin{figure}[t]
    \centering
        \includegraphics[trim={0 68px 0 0},clip,width=0.45\textwidth]{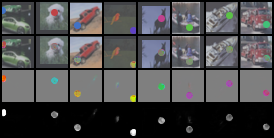}
    \caption{Examples of shortcuts and lens output on CIFAR10 training data.
    \textbf{Row 1} shows the input image with the color dot shortcut added:
    for example, all cars have a blue dot.
    \textbf{Row 2} shows the output of the lens, where the shortcut is mitigated by recoloring.
    }
    \label{fig:cifar_result}
\end{figure}

\textbf{Visualization}: \Cref{fig:cifar_result} presents example outputs of the attention lens when training on the CIFAR10 Color Dot shortcut. We make the following observations based on the visualization: Firstly, the attention lens successfully removes the shortcuts from the image. Secondly, for the Color Dot shortcut, recoloring the dots is sufficient to eliminate the shortcut as only the color of the dot is deterministic of the class. Additionally, we perform similar experiments for the Location Dot shortcut.
The model correctly learns that the Location Dot shortcut cannot be removed by recoloring the dots. Instead, the lens fills the dots by in-painting a best-effort background. 

\begin{figure}
    \centering
    \resizebox{.4\textwidth}{!}{%
    \begin{tikzpicture}
        \begin{axis}[
            xlabel={$\rho$},
            ylabel={accuracy},
            xticklabel={\pgfmathparse{\tick*100}\pgfmathprintnumber{\pgfmathresult}\%},
            yticklabel={\pgfmathparse{\tick*100}\pgfmathprintnumber{\pgfmathresult}\%},
            xmin=-0.02,
            ymax=0.8,
            ymin=0.2,
            width=8cm,
            height=4.7cm,
            ytick distance=0.2,
            minor x tick num=1,
            ymajorgrids,
            xmajorgrids
        ]
            \addplot [
              black, mark options={black, scale=0.75},
              only marks,
              error bars/.cd, 
                y fixed,
                y dir=both, 
                y explicit
            ]
            table [x=rho, y=mean, y error=error, col sep=semicolon] {figs/cifar_rhos.csv};
        \end{axis}
    \end{tikzpicture}
    }
    \caption{Accuracy on the \emph{clean} validation set when training our model on the CIFAR10 dataset, with varying degrees of $\rho$ (Location Dot shortcut). 
    }
    \label{fig:rho_graph}
\end{figure}
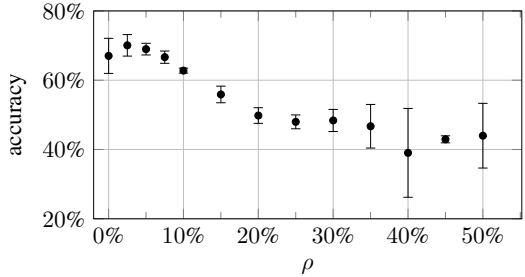

In order to determine the optimal value of $\rho$, we conduct a CIFAR10 Location Dot experiment with varying values of $\rho$. Specifically, we evaluate each of the candidate values for $\rho$ over three independent runs, and reported the mean accuracy and $95\%$ confidence interval in \cref{fig:rho_graph}. Our findings suggest that the optimal value of $\rho$ for this particular shortcut is around $\rho=2.5\%$, which approximately corresponds to the percentage of the image occupied by the shortcut. A significantly higher value of $\rho$ leads to the lens over-manipulating the image, resulting in a poor classifier performance on the original images.

\subsection{Real-World Data}
\begin{figure}[t]
    \centering
    \begin{subfigure}[b]{0.145\textwidth}
        \includegraphics[width=0.99\textwidth]{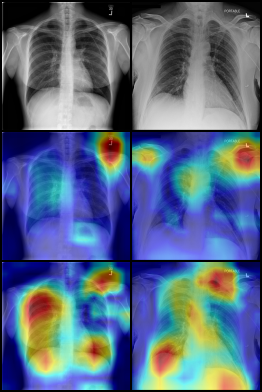}
        \caption{Normal}
    \end{subfigure}
    \begin{subfigure}[b]{0.145\textwidth}
        \includegraphics[width=0.99\textwidth]{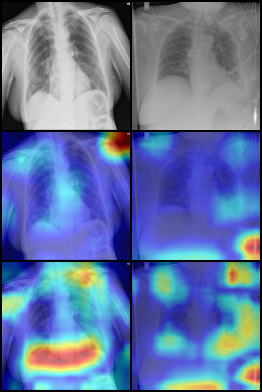}
        \caption{COVID}
    \end{subfigure}
    \begin{subfigure}[b]{0.145\textwidth}
        \includegraphics[width=0.99\textwidth]{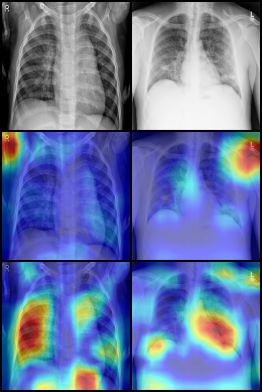}
        \caption{Pneumonia}
    \end{subfigure}

    \caption{GradCAM images showing network attention when training on the covid-qu-Ex dataset.
    \textbf{Row 1} is the input image from the validation set.
    \textbf{Row 2} is the classifier attention of a network trained without, and
    \textbf{Row 3} with our proposed model.}
    \label{fig:COVID_cam}
\end{figure}

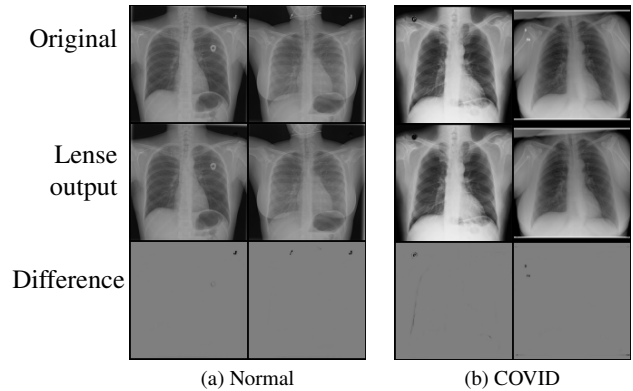
\begin{figure}[t]
    \centering
    \begin{minipage}[t]{0.08\textwidth}
        \begin{tikzpicture}
            \node [align=right, text width=\textwidth] at (0,77px) {Original};
            \node [align=right, text width=\textwidth] at (0,27px) {Lense output};
            \node [align=right, text width=\textwidth] at (0,-13px) {Difference};
            \node [align=right, text width=\textwidth] at (0,-40px) {};
        \end{tikzpicture}
    \end{minipage}
    \hspace{0.1cm}
    \begin{subfigure}[t]{0.18\textwidth}
        \includegraphics[trim={130px 130px 0 0},clip,width=\textwidth]{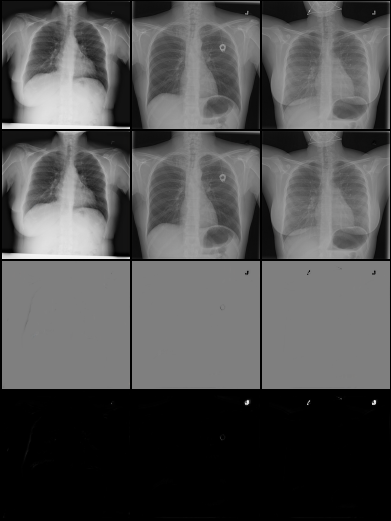}
        \caption{Normal}
    \end{subfigure}
    \hfill
    \begin{subfigure}[t]{0.18\textwidth}
        \includegraphics[trim={130px 130px 0 0},clip,width=\textwidth]{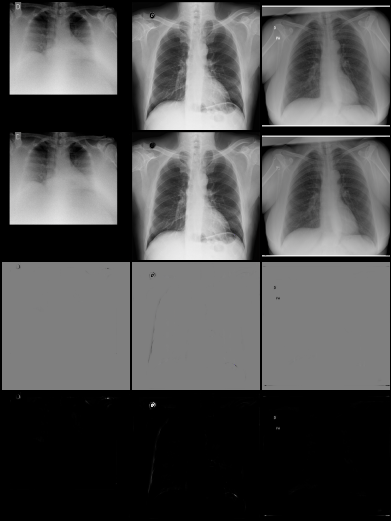}
        \caption{COVID}
    \end{subfigure}
    \caption{Lens output and attention on x-ray images from the covid-qu-Ex dataset for the classes classes COVID and Normal.
    \textbf{Row 1} shows original images.
    \textbf{Row 2} shows the output of the lens.
    \textbf{Row 3} shows the difference between rows 1 and 2.
    }
    \label{fig:COVID_result}
\end{figure}

For the covid-qu-Ex dataset, we trained the network with hyperparameters $\lambda=5$, $\rho=0.25\%$, 2 downsampling steps in the attention network, and 5 downsampling steps in the replacement network. We used a learning rate of $2 \cdot 10^{-4}$ for both the lens and classifier.
As there is no validation set without shortcuts for covid-qu-Ex, we evaluated the effectiveness of the lens in identifying shortcuts using GradCAM \cite{Selvaraju_2017_ICCV}. \Cref{fig:COVID_cam} shows the GradCAM images for all three classes and both trained networks. From these experiments, we made several observations. First, without the lens, the network predominantly focused on areas in the corners of the images, mostly in areas with text. Second, with the attention lens, the network focused on more relevant sections of the image, including the lungs.
Our proposed approach not only explains shortcuts but also corrects them, as shown in \cref{fig:COVID_result}, where highly localized shortcuts such as markers and text are removed.

\section{Conclusion}
In this paper, we propose a method for detecting and eliminating small but highly influential shortcuts in machine learning datasets. 
Our approach is built upon the hypothesis that genuine features are typically more global in nature, whereas shortcuts are localized but highly predictive.
However, we acknowledge that there may be datasets containing global shortcuts such as image background~\cite{sagawa2019distributionally} or ambient lighting, but leave this for future work.
To validate our proposed approach for localized shortcut detection, we conduct experiments on both synthetic and real-world datasets and demonstrate our model's effectiveness.

{\small
\bibliographystyle{ieee_fullname}
\bibliography{references}
}

\end{document}